\documentclass{article}

     \usepackage[final]{neurips_2019}

\usepackage[utf8]{inputenc} % allow utf-8 input
\usepackage[T1]{fontenc}    % use 8-bit T1 fonts
\usepackage{hyperref}       % hyperlinks
\usepackage{url}            % simple URL typesetting
\usepackage{booktabs}       % professional-quality tables
\usepackage{amsfonts}       % blackboard math symbols
\usepackage{nicefrac}       % compact symbols for 1/2, etc.
\usepackage{microtype}      % microtypography
\usepackage{xcolor}
\usepackage{hyperref}
\hypersetup{
    colorlinks,
    linkcolor={red!50!black},
    citecolor={blue!50!black},
    urlcolor={blue!80!black}
}

\title{Biology and Compositionality: Empirical Considerations for Emergent-Communication Protocols}

\author{
  Travis~LaCroix\thanks{\href{travislacroix.github.io}{\texttt{http://travislacroix.github.io}}} \\
    \phantom{AND} \\
  Department of Logic \& Philosophy of Science\\
  University of California, Irvine\\
  Irvine, CA \\
  \phantom{AND} \\
  Mila \\
  Qu{\'e}bec AI Insitute \\
  Montr{\'e}al, PQ \\
  \phantom{AND} \\
  \texttt{tlacroix@uci.edu} \\
}

\begin{document}

\maketitle

\begin{abstract}
    Many researchers in language origins and emergent communication take {\it compositionality} as their primary target for explaining how simple communication systems can become more like natural language. % %Demonstrating how compositionality may have evolved is supposed to be (at least partially) sufficient for explaining how language evolved.
    I suggest that, if machine learning research in emergent communication is to take cues from biological systems and language origins, then compositionality is not the correct target.
\end{abstract}

\section{Introduction}

Communication is ubiquitous in nature, but {\it linguistic} communication---i.e., natural language---is unique to humans. There are inherent difficulties in determining how language may have evolved---i.e., out of simpler communicative precursors ({\it proto-language}). On the one hand, it is difficult to define what constitutes language---it is always in flux, it is infinitely flexible, and it is one of the most complex behaviours known \citep{Christiansen-Kirby-2003}. On the other hand, there is a paucity of direct evidence available with which to confirm current theories---language does not fossilise, nor can we go back in time to observe the actual precursors of human-level linguistic capacities.

Language origins research may appear to be a purely epistemic project; however, this programme finds application in contemporary machine learning and artificial intelligence (ML/AI). Understanding the fundamental principles that are involved in the (biological and cultural) evolution of effective communication may lead to innovative communication methods for use by interacting AI agents and multi-robot systems \citep{Wagner-et-al-2003}. AI agents need a common language to coordinate successfully and to communicate with humans. Furthermore, given the close relationship between language and thought, a clear understanding of how languages arise and persist in a population may also lend some additional insight into human cognitive capacities, helping lead the way in creating an artificial {\it general} intelligence (AGI).

The similar goals of language origins research (from a philosophical, biological, bio-linguistic, and cognitive systems perspective) and emergent communication research (from a computational perspective) should be apparent. One of the questions asked in contemporary machine learning is
\begin{center}
    {\it How can emergent communication become more like natural language?}
\end{center} 
This question finds parallel expression in language origins research:
\begin{center}
    {\it How might something like natural language evolve out of a simple communication system?}
\end{center}
Thus, progress in language origins may help direct research in emergent communication and vice-versa. However, there has been relatively little dialogue between these two approaches.

To explain how language evolved out of simpler precursors, it is necessary to understand the salient differences between language and simple communication, or signalling. Most researchers (in both programmes) working on these questions take {\it compositionality} (productivity, openness, hierarchical structure, generative capacity, complex syntax, etc.) as their primary target. Demonstrating how compositionality may have evolved is supposed to be (at least partially) sufficient for explaining how language evolved. Furthermore, narrowing one's explanatory target from the formidable question of how {\it language} evolved to the question of how one component of language---e.g.,  {\it complex syntax}---evolved affords some degree of analytic tractability. 

However, there is reason to think that compositionality is the wrong target, on the biological side (and so the wrong target on the ML side). As such, the purpose of this paper is to explore this claim. %, %
%the titular conditional statement, 
%focusing upon the consequences for potential research in emergent communication. %As a result, I will not argue about {\it whether} gradualism is the correct approach to language origins research; this will be beside the point since, of course, one cannot deny a conditional statement by denying the antecedent. Thus, I will assume that the antecedent is true and then proceed to discuss why the consequent follows from this assumption. 
This has theoretical implications for language origins research more generally, but the focus here will be the % significant 
implications for research on {\it emergent communication} in computer science and ML---specifically regarding the types of programmes that might be expected to work, and those which will not. My hope, then, is that this work will help to direct future research in fruitful new directions. 

\section{A Quick Note on Compositionality}

Compositionality provides a {\it prima facie} plausible explanatory target for language origins and emergent communication research. In the former case, this is because compositional syntax is a salient differentiating feature of language which is absent in simple communication systems found in nature. The {\it principle} of compositionality is typically stated as follows: {\it The meaning of a compound [complex] expression is a function of the meaning of its parts [constituents] and how they are combined [composed]} \citep{Kamp-Partee-1995, Szabo-2012}. In ML/AI, this is usually cashed out as a notion of {\it systematicity}; namely, ``the ability to entertain a given thought implies the ability to entertain thoughts with semantically related contents'' \citep[3]{Fodor-Pylyshyn-1988}. 

Any attempt to give an evolutionary explanation of human-level linguistic capacities will minimally need to account for the following: (1) how compositionality might arise from non-compositional communication; (2) if compositionality itself is an evolutionary adaptation, why compositional structure should be selected for in the first place; (3) why compositionality should be rare in nature, though communication is universal. %
Several models\footnote{See \citet{Nowak-Krakauer-1999, Barrett-2006, Barrett-2007, Barrett-2009, Franke-2014, Franke-2016, Steinert-Threlkeld-2014, Steinert-Threlkeld-2016, Steinert-Threlkeld-2018, Barrett-et-al-2018}.} have been suggested in recent years which grapple with these questions using the signalling-game framework.\footnote{The signalling game was introduced in \citet{Lewis-1969} and extended to dynamic models in \citet{Skyrms-1996, Skyrms-2010-Signals}. In the ML literature, this is sometimes called a {\it referential game}; see \citet{Lazaridou-et-al-2016, Das-et-al-2017, Evtimova-et-al-2017, Havrylov-Titov-2017, Kottur-et-al-2017, Choi-et-al-2018, Lazaridou-et-al-2018}.} Signalling games show how straightforward communication conventions might arise naturally through processes of repeated interactions. In an evolutionary context, starting with initially random signals and actions, individuals in a population learn or evolve effective communication conventions under several different dynamics. 

However, \citet{LaCroix-2019-Compositionality} suggests that the evolutionary explanations for compositional {\it signalling} offered thus far fail to give a plausible account of how compositionality might arise. The reason for this failure is twofold. On the one hand, these models often (if implicitly) take compositionality {\it qua} linguistic compositionality as their target for an evolutionary explanation. This gives rise to significant complications insofar as linguistic compositionality is rife with conceptual difficulties. On the other hand, these models fail to take into account the role asymmetry of the sender and receiver in the signalling game and thus fail to capture how compositionality might be beneficial for communication. \citet{LaCroix-2019-Compositionality} suggests that if compositionality is a good target of language origins (or emergent communication) research, then it is necessary to build a more straightforward notion of  compositionality for complex signalling from the bottom-up. Here, I suggest that compositionality is not even the right target for these research programmes. 

\section{Language Origins}

In the context of the salient distinctions between language and animal communication systems, the problem is that compositional signals are extremely rare, or nonexistent, in nature. There is scant empirical evidence for call sequences that are {\it compositionally} meaningful. Most current data that suggests compositional signalling in nature comes from Zuberb{\" u}hler's study of several species of African forest monkeys ({\it Cercopithecus}) \citep{Zuberbuhler-2001, Zuberbuhler-2002, Arnold-Zuberbuhler-2006a, Arnold-Zuberbuhler-2006b, Arnold-Zuberbuhler-2008, Arnold-Zuberbuhler-2013}. However, no such combinatorial capacity or anything similar has been found in {\it any} other species of monkey or great ape, though these are well-studied \citep{Fitch-2010}.

As a result, there is little evidence of a precursor to human-level compositional syntax. The most recent last common ancestor (LCA) between the great apes (including humans) and old-world monkeys (including {\it Cercopithecus}) existed approximately $25$ {\sc mya} (million years ago), during the Paleogene period. Since no other related species utilises such compositional capacities, it stands to reason that this syntactic disposition was {\it not} present in the LCA of {\it Homo} and {\it Cercopithecus}---the most recent common ancestor of humans and chimpanzees, for example, lived around $10-4$ {\sc mya}. Australopithecine species evolved after this break, with {\it Homo} likely evolving approximately $2$ {\sc mya}. 

Suppose that the LCA to great apes and old-world monkeys {\it did} have functionally proto-compositional syntax. This requires that proto-compositional syntax evolved up to $25$ {\sc mya}. Then, we should expect that most old-world monkeys {\it and} great apes would show signs of using proto-compositional syntax. However, as far as we can tell, they do not---excepting the few mentioned {\it Cercopithecus} species. Assuming these dispositions evolved in the LCA, $25$ {\sc mya}, and given that they do not appear in most decedents of the LCA, this implies that this disposition was lost---and lost more readily than it was held onto---in almost all other {\it Catarrhine} species.

Further still, it is controversial whether Neanderthals had language ($0.4-0.04$ {\sc mya}), let alone whether {\it H. heidelbergensis} had language ($0.7-0.2$ {\sc mya}). But, it is generally accepted that {\it Australopithecine} species ($3.9-2.9$ {\sc mya}) did {\it not} have language \citep{Fitch-2010}. This implies that this proto-compositional disposition evolved into fully-fledged natural language on the order of $0.2-2.0$ {\sc mya}.\footnote{\citet{Berwick-Chomsky-2016} give a more conservative estimate, between $0.06-0.2$ {\sc mya}, corresponding to the first anatomically-modern humans and the last exodus from Africa.} Meanwhile, these dispositions in few {\it Catarrhine} species remained utterly unchanged for more than $20$ million years, and these dispositions were lost in {\it almost all} great apes and old-world monkeys. This is the implausible (though technically not impossible) conclusion that follows from our assumption that a proto-compositional disposition existed in the LCA of great apes and old-world monkeys.  

Thus, the scant empirical evidence for compositional precursors to human language comes paired with an improbable evolutionary history. The more plausible alternative is that there is {\it no} empirical evidence for any precursor to human language in nature. Note that an account which posits the sudden emergence of compositional syntax does not fall prey to these criticisms since it requires no proto-compositional precursor to compositional language. This is a virtue of the so-called {\it saltationist} perspective on language origins. However, this view is not without its own problems \citep{LaCroix-2019-Salt-v-Grad}. Therefore, {\it if} gradualism is the correct approach to language origins, then compositionality {\it cannot} be a correct target.

\section{Consequences for Machine Learning}

Significant advances have been made in artificial systems by using biological systems as a guide. For example, the development of reinforcement learning algorithms was heavily inspired by learning rules that were studied empirically in biology \citep{Bush-Mosteller-1955, Rescorla-Wagner-1972, Roth-Erev-1995, Erev-Roth-1998}. A recent review of points of contact between ML and biology suggests several different areas of future research which can benefit from a bidirectional flow of information between these fields \citep{Neftci-Averbeck-2019}; however, nowhere is mentioned the interface between biological and computational models of the emergence of language.

I suggested above that a gradualist approach to language origins in biology implies that compositionality is an incorrect target for this research programme. Further, computer scientists working on emergent communication also take compositionality as a primary target---a benchmark for success. However, recent work in ML highlights several problems for learning compositional linguistic structures. In particular, neural networks latch on to statistical regularities in existing datasets: \citet{Bahdanau-et-al-2018} highlight that, in a synthetic instruction-following task \citep{Lake-Baroni-2017}, the agent does not learn a generalisation for composing words. Thus, when the agent is trained on the commands `jump', `run twice', and `walk twice', it subsequently fails when asked to interpret `jump twice'. Thus, they fail to {\it generalise}. %
%
%Neural networks are the `workhorse' of natural language comprehension and generation---\citet{Bahdanau-et-al-2018} highlight that neural networks play a significant role in machine translation \citep{Wu-et-al-2016} and text generation \citep{Kannan-et-al-2016} in addition to exhibiting state-of-the-art performance on several benchmarks, including {\it Recognising Textual Entailment}, \citep{Gong-et-al-2017}, {\it Visual Question Answering}, \citep{Jiang-et-al-2018}, and {\it Reading Comprehension} \citep{Wang-et-al-2018}. However, they 
%
The above analysis suggests a possible reason why they do so fail: the target is compositional communication. Thus, if ML is to take cues from biological systems, it seems that this is not the correct direction for research in emergent communication.

\section{Where to Go From Here: Reflexivity}

Communication is a unique evolutionary process in the following sense: once a group of individuals has learned some set of simple communication conventions, those learned behaviours may be used to influence future communicative behaviours, giving rise to a feedback loop. When faced with a novel context, an individual can always learn a brand new disposition. However, the individual may learn to take advantage of previously evolved dispositions. Indeed, they may learn to take advantage of pre-evolved {\it communicative} dispositions to thereby influence the evolution of future communicative dispositions. This is a notion of {\it reflexivity}. Reflexivity, unlike compositionality, is consistent with a gradualist approach to language origins. %
Once actors exhibit such complexity, at a small scale, it may lead to a feedback loop between communication and cognition that, over time, gives rise to the complexity that we see in natural languages. Thus, this evolutionary story depends inherently on a notion of {\it conceptual bootstrapping} \citep{Carey-2004, Carey-2009a, Carey-2009b, Carey-2011b, Carey-2011a, Carey-2014, Shea-2009, Margolis-Laurence-2008, Margolis-Laurence-2011, Piantadosi-et-al-2012, Beck-2017}.

To improve performance on generalisation, researchers in ML might add modularity and structure to their designs \citep{Andreas-et-al-2016, Gaunt-et-al-2016}. In the case of the Neural Module Network paradigm, a neural network is assembled from several {\it modules}, each of which is supposed to perform a particular subtask of the input processing. \citet{Bahdanau-et-al-2018} note that although this modular approach is intuitively appealing, ``widespread adoption has been hindered by the large amount of domain knowledge that is required to decide or predict how the modules should be created \ldots and how they should be connected \ldots based on a natural language utterance''. See also, \citet{Andreas-et-al-2016, Johnson-et-al-2016, Johnson-et-al-2017, Hu-et-al-2017}. %
Insofar as reflexivity is an apt target for the biological evolution of linguistic communication, it may too provide some insights for modelling emergent communication in an artificial system. See further discussion in \citet{LaCroix-If-Gradualism-2019, LaCroix-Dissertation}.

\section{Further Resources}

The gradualist view (though not necessarily via natural selection) is endorsed by, e.g., \citet{ %
    Givon-1979, 
        Givon-2002a, 
        Givon-2002b, 
        Givon-2009, 
    Pinker-Bloom-1990, 
    Newmeyer-1991, 
        Newmeyer-1998, 
        Newmeyer-2005, 
    Jackendoff-1999, 
        Jackendoff-2002, 
    Carstairs-McCarthy-1999, 
    Fitch-2004, 
        Fitch-2010, 
    Culicover-Jackendoff-2005, 
    Szamado-Szathmary-2006,
    Progovac-2006, 
        Progovac-2009a, 
        Progovac-2009b, 
        Progovac-2013, 
        Progovac-2015, 
        Progovac-2019, 
    Tallerman-2007, 
        Tallerman-2013a, 
        Tallerman-2013b, 
        Tallerman-2014a, 
        Tallerman-2014b,
    Heine-Kuteva-2007, 
    Hurford-2007, 
        Hurford-2012,
    Tallerman-Gibson-2011,
    Yang-2013}, among many others.
The saltationist view is endorsed by \citet{%
    Berwick-1998, 
        Berwick-et-al-2013, 
        Berwick-Hauser-Tattersall-2013, 
    Bickerton-1990, 
        Bickerton-1998, 
    Lightfoot-1991, 
    Hauser-et-al-2002, 
    Chomsky-2002, 
        Chomsky-2005, 
        Chomsky-2010, 
    Piattelli-Palmarini-Uriagereka-2004,
        Piattelli-Palmarini-Uriagereka-2011, 
    Moro-2008, 
    Hornstein-2008, 
    Piattelli-Palmarini-2010, 
    Berwick-Chomsky-2011, 
        Berwick-Chomsky-2016, 
    Di-Sciullo-2011, 
        Di-Sciullo-2013, 
    Bolhuis-et-al-2014, 
    Miyagawa-et-al-2014, 
        Miyagawa-2017}, etc.

%I have not argued here for the virtues of the former approach over the latter; though see \citet{LaCroix-2019-Salt-v-Grad}.

This is to say nothing of the signalling-game framework \citep{Lewis-1969, Skyrms-1996, Skyrms-2010-Signals} in evolutionary game theory, which has seen a number of significant advances in a variety of philosophically interesting domains. These include, e.g., the difference between indicatives and imperatives \citep{Huttegger-2007b, Zollman-2011}; signalling in social dilemmas \citep{Wagner-2014}; network formation \citep{Pemantle-Skyrms-2004, Barrett-et-al-2017}; deception \citep{Zollman-et-al-2013, Martinez-2015, Skyrms-Barrett-2018}; meta-linguistic notions of truth and probability \citep{Barrett-2016, Barrett-2017}; syntactic structure and compositionality \citep{Franke-2016, Steinert-Threlkeld-2016, Barrett-et-al-2018, LaCroix-2019-Logic-Game}; vagueness \citep{OConnor-2014}; and epistemic representations, such as how the structure of one’s language evolves to maintain sensitivity to the structure of the world \citep{Barrett-LaCroix-2019}. See \citet{LaCroix-2019-Signaling-Models} for an overview.

\small
\bibliographystyle{apalikelike}
\bibliography{Biblio}

\end{document}